\documentclass[conference]{IEEEtran}
\IEEEoverridecommandlockouts
\usepackage{cite}
\usepackage{amsmath,amssymb,amsfonts}
\usepackage{algorithmic}
\usepackage{graphicx}
\usepackage{textcomp}
\usepackage{subcaption}
\usepackage{multirow}
\usepackage{bbm}
\usepackage{xcolor}
\usepackage[hyphens]{url}
\usepackage{diagbox}

\begin{document}

\title{Learning Task-Independent Game State Representations from Unlabeled Images \\
% Learning General Game State Representations \\ from Unlabeled Images \\
%\thanks{Chintan Trivedi, Antonios Liapis and Georgios N. Yannakakis were supported by the European Union’s H2020 research and innovation programme (Grant Agreement No. 951911). Konstantinos Makantasis was supported by the European Union’s H2020 research and innovation programme (Grant Agreement No. 101003397).}}
\thanks{This project has received funding from the EU’s Horizon 2020 research and innovation programme under grant agreements No 951911 and No 101003397.}}

\author{\IEEEauthorblockN{Chintan Trivedi,
Konstantinos Makantasis, Antonios Liapis and
Georgios N. Yannakakis}
\IEEEauthorblockA{Institute of Digital Games,
University of Malta, 
Msida, Malta\\
Email: 
%\IEEEauthorrefmark{1}ctriv01@um.edu.mt,
%\IEEEauthorrefmark{2}\{firstname\}.\{lastname\}@um.edu.mt}}
\{ctriv01, konstantinos.makantasis, antonios.liapis, georgios.yannakakis\}@um.edu.mt
}}
\maketitle

\begin{abstract}
Self-supervised learning (SSL) techniques have been widely used to learn compact and informative representations from high-dimensional complex data. In many computer vision tasks, such as image classification, such methods achieve state-of-the-art results that surpass supervised learning approaches. In this paper, we investigate whether SSL methods can be leveraged for the task of learning accurate state representations of games, and if so, to what extent. For this purpose, we collect game footage frames and corresponding sequences of games' internal state from three different 3D games: VizDoom, the CARLA racing simulator and the Google Research Football Environment. We train an image encoder with three widely used SSL algorithms using solely the raw frames, and then attempt to recover the internal state variables from the learned representations. Our results across all three games showcase significantly higher correlation between SSL representations and the game's internal state compared to pre-trained baseline models such as ImageNet. Such findings suggest that SSL-based visual encoders can yield \emph{general}---not tailored to a specific task---yet \emph{informative} game representations solely from game pixel information. Such representations can, in turn, form the basis for boosting the performance of downstream learning tasks in games, including gameplaying, content generation and player modeling.
\end{abstract}


\section{Introduction}\label{sec:intro}
\begin{figure}[t]
\centering
\includegraphics[width=\columnwidth]{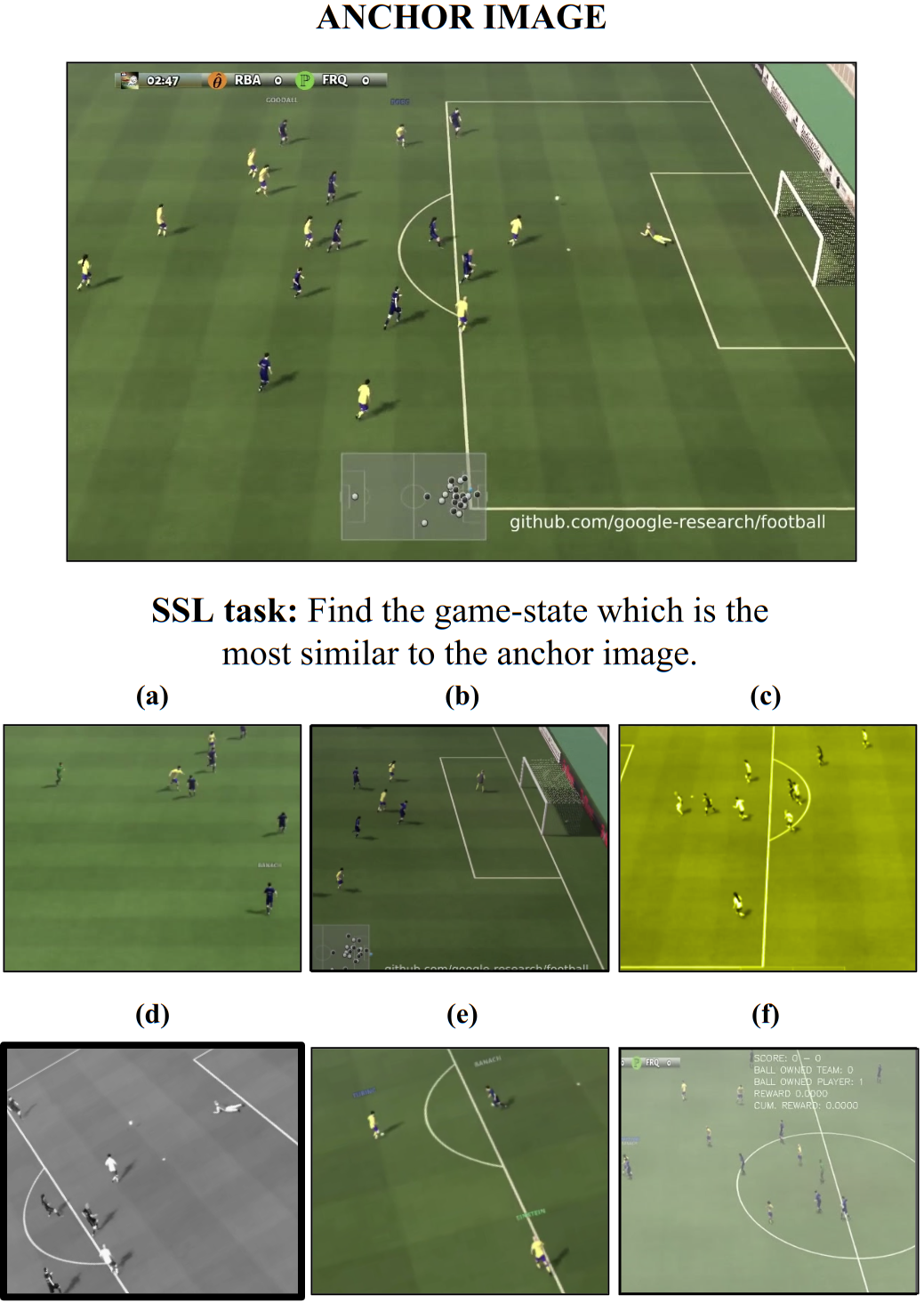}
\caption{Illustrative example of a Self Supervised Learning task that facilitates learning of visual features of football and is able to distinguish between different states of the game.}
\label{fig:sslexample}
\end{figure}

\begin{figure*}[!tb]
\centerline{\includegraphics[width=\textwidth]{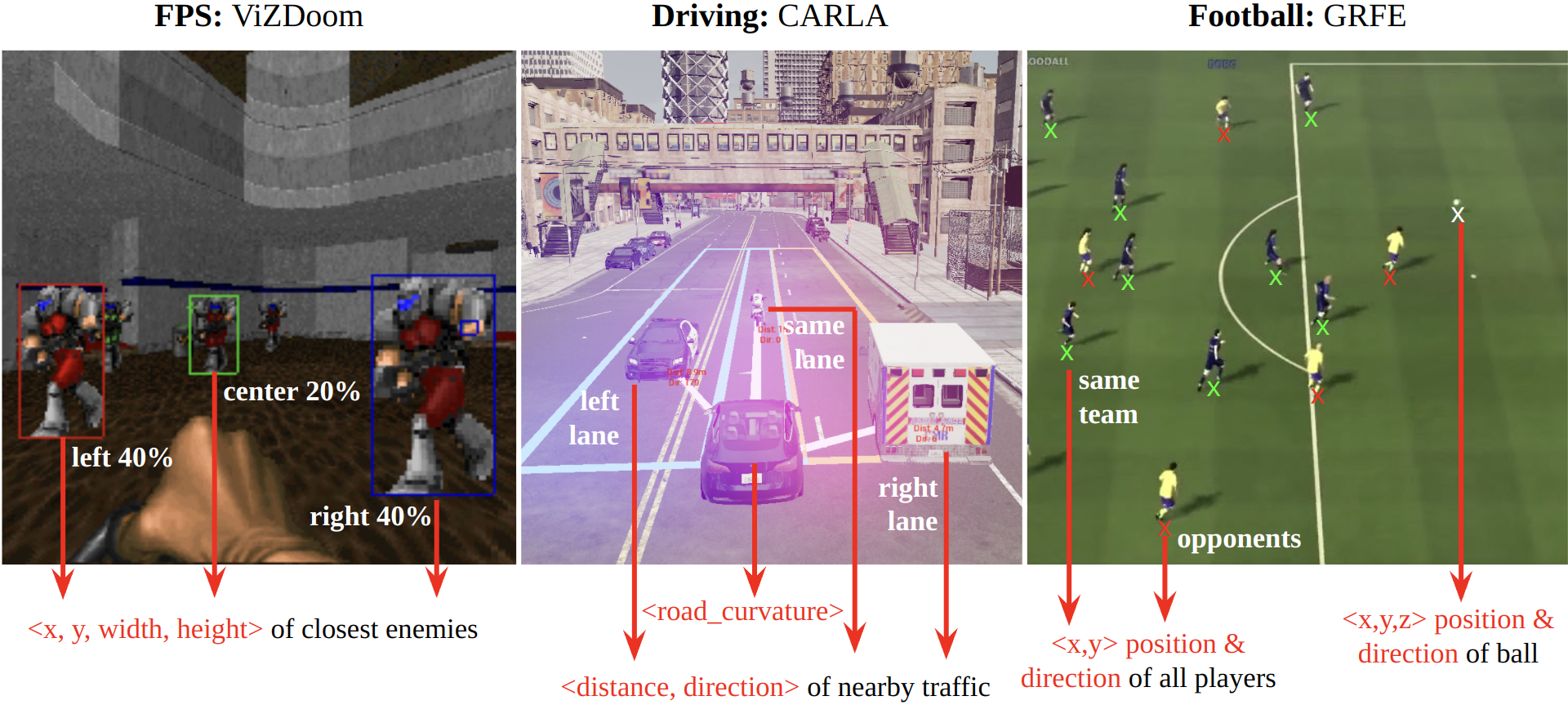}}
\caption{Games and their \emph{critical} game-state features included in the 3D-SSL Benchmark Dataset used in our experiments.}
%\caption{The three game environments used in our experiments. All three provide access to their game engines and hence, we are able to extract accurate and precise internal state values associated with each frame. The features (internal state variables) illustrated for each game are hand-engineered such that the values obtained from the game engine are either clearly visible or can be easily inferred from the image. Note that the three games are rather representative and differ in terms of genre (i.e. FPS, racing, sports), image resolution (i.e. photorealistic vs. pixelated) and key information depicted such as size and amount of objects which are key to the game.}
\label{fig:dataset}
\end{figure*}

Research on representation learning within the field of computer vision aims to transform high-dimensional pixel information to compact low-dimensional vector embeddings that capture the essential features describing the image content. These compact representations are assumed to be general---i.e. not tailored to a specific task---and can be used in any vision-based learning task such as object recognition, object detection and image segmentation. Representation learning in digital games, however, remains an open challenge \cite{yannakakis2018artificial}. Many end-application tasks in games research---including gameplaying, modeling user behavior, and content generation---use image representations of games containing information about \emph{critical} factors describing the current state of the game \cite{anand2019unsupervised}. These critical factors are defined in relation to the specific objectives and constraints of the particular game world (see Fig. \ref{fig:dataset} for examples). Along with the presence and identity of objects in a gameplay frame, a good representation should be able to ignore the game's aesthetics \cite{trivedi2021contrastive}, capture the spatio-temporal relations among objects---even if occluded or off-screen---and infer the dynamics and underlying rules of the game \cite{ha2018world}. 
% A number of end-application tasks in games research including gameplaying, modeling user behavior, and generation of game content use image representations of games describing the critical factors of the game's state. We define these \emph{critical} (or informative) factors of any game state as the description of any object in the game-world at a moment in time.
% \textcolor{red}{In other words, it %Such representations 
% should capture game's factors 
%the essential latent generative factors 
% \cite{anand2019unsupervised, khan2021pretrained} 
%of the game 
% that describe game-play in terms of defined rules and constraints of the game world.}
%Therefore, game representation learning in games focuses mainly on extracting the generative factors and variables that determine the rendering result corresponding to the frames of the game. 
Having access to the game model---and thus, the game's internal state---renders representation learning trivial or even redundant, as visual information can be explicitly mapped to the internal state. Game models, however, are rarely available (especially for commercial games) and thus representation learning is required to implicitly map the visual information obtained from a game's footage to the game's internal state. 

In this study, we investigate the degree to which we can learn the mapping from gameplay frames to the game's model---as represented by critical internal variables---without any knowledge about games' internal state. For this purpose, we take advantage of popular self-supervised learning (SSL) algorithms widely used for learning general representations in a plethora of computer vision tasks \cite{zbontar2021barlow,chen2020exploring,bardes2021vicreg}. SSL refers to a family of machine learning algorithms that can learn essential features describing the visual content of images without requiring any label information, i.e. image characterization \cite{chen2020simple}. SSL algorithms thus alleviate the tedious and time consuming process of data labelling. Instead, training is based on the inherent properties and characteristics of the images' visual content, which are used as learning signals. 

To better realise how SSL can be utilised for state representation learning in games, consider the example of an auxiliary task shown in \figurename{} \ref{fig:sslexample}. An anchor image is presented along with six different states of the same football game. The objective is to find which of the six images corresponds to the game \textit{state} depicted in the anchor image. To achieve that objective, one needs to be able to extract the critical gameplay factors from the images of the game, such as the relative position of the players and the ball on the football pitch and disregard non-critical factors related to game aesthetics (e.g. ambient light levels or the advertisements displayed beyond the pitch).
%\textcolor{red}{An anchor image is presented along with six different states of the same football game. Among these, only one corresponds to the game state depicted in the anchor image and one is tasked to find that image. Here, the \emph{state} of the game of football is defined by the critical gameplay factors like the relative position of the players and the ball on the football pitch, whereas the aesthetics of the pitch or changes in advertisement boards can be considered as the non-critical factors. Thus, to solve this task, one needs to understand and extract the critical gameplay factors from the images of the game.} 
Based on such information, one can deduce that \textit{(d)} is the view that most closely matches the anchor image. Thus, in trying to solve this auxiliary task, one ends up learning to predict the important features of a football game---i.e. the spatial information regarding the positions of players and the ball---from just the image of the game. This is how an SSL algorithm operates and learns general representations from a high-level perspective, which can, in turn, be used for any downstream task that requires such a representation of the game state.

This paper introduces the notion of SSL within games and explores its capacity for state representation learning in 3D games. Our hypothesis is that SSL is beneficial for learning representations that are able to infer core elements of games, their objects and their underlying mechanics without access to such data. To test our hypothesis we employ three popular SSL methods with dissimilar learning properties---SimCLR \cite{chen2020simple}, SwAV \cite{caron2020unsupervised} and BYOL \cite{grill2020bootstrap}---and test them across three games that vary with regards to their genre, image resolution, artistic style and object properties (i.e. size and number of objects). In particular we train SSL methods on a new benchmark dataset we name 3D-SSL that contains $150,000$ game footage frames of VizDoom \cite{kempka2016vizdoom}, CARLA driving simulator \cite{dosovitskiy2017carla} and Google Research Football Environment \cite{kurach2019google} and we evaluate the models' capacity to predict internal game variables (e.g. the position of the ball, teammates, enemies or cars) on over $10,000$ frames per game. Results suggest that SSL---compared to pretrained models such as ImageNet---learn general representations that are able to predict the internal state of all 3D games with high degrees of $R^2$ correlation via \emph{linear probing} \cite{anand2019unsupervised}. The key findings of the paper indicate that SSL is a highly recommended method for constructing game state representations that can be employed for any downstream task that requires such a game state representation including gameplaying, content generation or player modeling \cite{yannakakis2018artificial}.

%The rest of the paper is organized as follows: Section \ref{sec:background} gives an overview of the different approaches to representation learning in games and introduces the notion of using SSL for this task. Section \ref{sec:dataset} introduces the 3D games we use in our experiments and introduces the dataset for testing and bench-marking SSL methods on these games. In section \ref{sec:sslmethods} we explain the SSL methods we focus on in this paper. Section \ref{sec:method} describes the training and evaluation approaches of our experiments. Finally, we present our results in section \ref{sec:results} and describe potential use-cases of this work in section \ref{sec:discussion}.

\section{Background}\label{sec:background}

% \textcolor{red}{I suggest to split this section into two subsections: the first one should be about SSL in Games (including state representation learning), and the second about Pixel-based inference in games (e.g. affect modelling, player behaviour modelling, game play agent etc. In addition, the last paragraph of this section should emphasize the novelty of this study--something such as ``our contribution''. )}

% Contrastive methods such as SwAV \cite{caron2020unsupervised}, Barlow Twins \cite{zbontar2021barlow}, BYOL \cite{grill2020bootstrap}, SimSiam \cite{chen2020exploring}, VICReg \cite{bardes2021vicreg}, DINO \cite{caron2021emerging}, SimCLR \cite{chen2020simple}.

% Deep Reinforcement Learning \cite{kirk2021survey}.
A substantial volume of AI and games research \cite{yannakakis2018artificial} focuses on the use of AI for building game-playing agents, for modeling players and their emotions, and for generating aspects of game content. Feeding directly the pixels of the game to the input of any AI model---predominately a neural network model---remained a challenging task for many of these applications owing to the high-dimensional nature of images. For that purpose, the majority of such studies have traditionally used some form of internal state representations of the game coming directly from interfacing with the game engine (e.g. \cite{barthet2021go, melhart2021affect, berner2019dota, silver2017mastering} among many). 

The recent success of Convolutional Neural Networks (ConvNets) in dimensionality-reduction for image processing has enabled research with raw game footage pixels. Indicatively, a number of recent studies have effectively used ConvNets with reinforcement learning for playing Atari games \cite{kirk2021survey,mnih2013playing,zhang2020atari} by mapping the game's raw image to an action to be performed for maximizing the game's score. Beyond gameplaying, a number of recent ConvNet studies focus on modeling players' affect \cite{makantasis2021privileged, makantasis2019pixels, makantasis2021pixels}. All of the aforementioned approaches, however, train ConvNets with task-specific output labels such as actions (i.e. for imitation learning), affect annotations (i.e. for affect modeling) or reward values (i.e. for reinforcement learning). This dependency on labelled data gives rise to two primary issues. First, while the labels are related to the learning task, they might not necessarily be informative about (or associated to) the visual information being processed \cite{stooke2021decoupling}. Second, these approaches tend to learn only highly task-specific information observed in the pixel input. 

As a response to these challenges, most modern approaches employing ConvNets attempt to separate the visual information processing part from the overall pipeline of the downstream application task. To this end, Chaplot \emph{et al.} \cite{chaplot2017arnold} used the game's internal state as additional labels within their reinforcement learning framework in order to guide the ConvNet to learn more useful visual features describing the game state. This approach, however, still requires access to the internal state of the game, making it impractical for most games where the world model is unavailable. A popular workaround is to simply use models pre-trained on large, universal image datasets such as ImageNet \cite{deng2009imagenet, poulsen2017dlne, trivedi2018building}, or massive generalized models such as CLIP \cite{radford2021learning, khan2021pretrained}. We use such approaches to form the baseline in our experiments and empirically showcase their shortcomings on capturing meaningful internal game states as depicted on game frames. 

In order to learn highly informative and compact state representations from images of the game, several self-supervised approaches have been used that learn using image-reconstruction techniques \cite{ha2018world, cao2017unsupervised, ledig2017photo}. The key advantage of these methods is that they do not require access to the game's internal world model. More recently, Anand \emph{et al.} \cite{anand2019unsupervised} proposed a self-supervised learning method which utilizes the spatial and temporal relations between frames of different Atari games to learn important visual features of the game's image. This approach, however, has two core limitations: first, it requires time-distributed images as the method's loss function incorporates temporal difference between the game's images and, second, it presents results on basic Atari games that are restricted to simple and abstract 2D grid environments, which are not representative of most modern era games.

% One such approach is to use an autoencoder to reconstruct the game image in the pixel space where the compressed encoding is expected to contain only the information useful for reconstructing the entire image \cite{ha2018world}. Similarly one may attempt to use ConvNet encoders from Generative Adversarial Networks (GANs) that are trained for image colorization \cite{cao2017unsupervised} or image super-resolution \cite{ledig2017photo}, assuming that the ConvNets would learn the important features of the game image while optimizing their respective learning objectives. Anand \emph{et al.} \cite{anand2019unsupervised} show, however, that such approaches are less effective compared to directly optimizing for learning the internal state variables. They

In this paper, instead, we attempt to address these two limitations by testing recent SSL methods that do not rely on images with any temporal association between them. Moreover, we extend SSL investigations to 3D games that provide more complex and challenging visual information to process. Importantly we test the capacity of such algorithms on capturing key internal game state variables across 3 very different games. We showcase that SSL, in contrast to pretrained image models, constructs general-purpose representations that can effectively predict such information.

\section{The 3D-SSL Benchmark Dataset} \label{sec:dataset}

As mentioned earlier, recent work in self-supervised representation learning considered 2D game environments such as Atari \cite{anand2019unsupervised}. Instead, this work investigates how these methods translate to more complex, 3D games with more sophisticated graphics and more detailed game states. Towards this endeavour, we choose three games from different game genres, representing different difficulty levels with regards to the task of obtaining an accurate game state from the image of the game. 
%
%AL: moving caption below
All three games provide access to their game engines and hence, we are able to extract accurate and precise internal state values associated with each frame. The features (internal state variables) illustrated for each game are hand-engineered such that the values obtained from the game engine are either clearly visible or can be easily inferred from the image. Note that the three games are rather representative and differ in terms of genre (i.e. FPS, racing, sports), image resolution (i.e. photorealistic vs. pixelated) and key information depicted such as size and number of objects.
In this section we outline the three games and the corresponding 3D-SSL Benchmark dataset, which is summarized in Table \ref{tab:dataset}.

\begin{table}[!tb]
\begin{center}
\caption{Summary of the 3D-SSL Benchmark Dataset.}
\label{tab:dataset}
\begin{tabular}{@{ }l@{ }||@{ }c@{ }|@{ }c@{ }|@{ }c@{ }}
\hline\hline

\textbf{\begin{tabular}[c]{@{}c@{}}\end{tabular}} & \textbf{\begin{tabular}[c]{@{}c@{}}ViZDoom\end{tabular}} & \textbf{\begin{tabular}[c]{@{}c@{}}CARLA\end{tabular}} & \textbf{\begin{tabular}[c]{@{}c@{}}GRFE\end{tabular}} \\ \hline \hline

Images (Training) & \ 50,000 \ & \ 50,000 \ & \ 50,000 \ \\ 
Image Size &  \ 400$\times$225$\times$3 \  & \ 224$\times$224$\times$3 \ & \ 224$\times$224$\times$3 \ \\
Image Frequency & \ 1 per time-step \ & \ 3 per second \ & \ 1 per time-step \ \\ 
Images (Evaluation) & \ 10,500 \ & \ 20,108 \ & \ 10,000 \  \\ 
State Variables & \ 12 \ & \ 7 \ & \ 94 \  \\ \hline\hline

\end{tabular}

\end{center}
\end{table}

\subsection{ViZDoom (First Person Shooter)}

From the shooter genre, we select the \emph{Doom} (Id Software, 1993) game via the ViZDoom environment \cite{kempka2016vizdoom}. We use a pre-trained model named ARNOLD \cite{chaplot2017arnold} to play the game and collect synchronized pairs of the RGB image of the game and its internal state. We represent the game's internal state using 12 features related to the enemy positions. In particular, we identify the closest enemy in the left $40\%$ of the screen, and use its $(x,y)$ location in screen buffer coordinates, and the width and height of its bounding box to represent its position on the screen. In the same manner we represent the position of the enemies in the middle $20\%$ and the right $40\%$ of the screen (see \figurename{} \ref{fig:dataset}). We choose to use these variables for representing the game's internal state since they largely determine the behaviour of the Doom player. In total, we collected $50,000$ images for training and $10,500$ pairs of images and the corresponding internal state variables for evaluation. ViZDoom represents a fairly easy task of state representation learning as the graphics of the game are low-resolution and all enemies have the same look and design. 

\subsection{CARLA (Racing)}

From the car racing genre, we use the CARLA open-source simulator for autonomous driving research \cite{dosovitskiy2017carla}. The inbuilt autopilot AI drives a car (\textit{ego-vehicle}) around an urban simulation environment, and we collect the game state information at 3 frames per second.
%to maintain a good degree of difference between the frames in terms of the information captured. 

The game state variables collected describe the nearby traffic for the ego-vehicle. To describe the traffic on the left lane of the ego-vehicle, we find the nearest vehicle that is in a region of 10 metres by 50 metres (see \figurename{} \ref{fig:dataset}) to the left of our vehicle and store its distance and direction. We do the same for the nearest vehicle on the right lane of the ego-vehicle as well as in front of the ego-vehicle. We also calculate the curvature of the road by measuring the angle between the direction of the ego-vehicle's current steering vector and the vector describing the curvature of the lane such that the vehicle remains at the center of the lane. The reason for choosing these 7 variables for describing the internal state (see \figurename{} \ref{fig:dataset}) is because they largely determine the steering and throttle inputs that should be given to the ego-vehicle in order to keep moving forward in the simulator environment. In total, we collected $50,000$ images for training and $20,108$ pairs of images with their corresponding internal state variables for evaluation. Here the evaluation set is larger than that of VizDoom because some variables such as traffic information have very low frequency, i.e., not every image has an associated variable value present in case there are no vehicles around our ego-vehicle.

CARLA presents a more challenging representation learning task than VizDoom: the graphics of the game are far more detailed, the vehicles surrounding the ego-vehicle have varying shapes and styling (unlike the enemies in VizDoom) and also determining the direction in which a vehicle is headed requires to acquire visual understanding of the design of these vehicles. 

\subsection{GRFE (Football)}

Lastly, we include the football simulator named Google Research Football Environment (GRFE) \cite{kurach2019google}. To create a training dataset, we trained a Proximal Policy Optimization \cite{schulman2017proximal} agent using stable-baselines \cite{raffin2021stable} to play an 11 vs 11 game against the inbuilt game bot and we collected RGB gameplay frames along with the game state at each timestep of the simulation.
The game state consists of the $(x,y)$ positions of each of the 22 players (11 per team) on the pitch along with the $(x,y)$ directions in which they are headed. Additionally, we collect the $(x,y,z)$ positions and directions of the ball. Positions are based on gameworld coordinates, rather than the screen buffer. The internal game state is represented by 94 variables (see \figurename{} \ref{fig:dataset}). We choose these variables since they are used as input by the PPO agent that plays this game and hence are considered as essential game state information that must be captured from the game's RGB gameplay frames. In total we collected $50,000$ training images and $10,000$ pairs of images and corresponding internal state values for evaluation.

GRFE is the most difficult state representation learning task since it involves inferring the precise location and movement information of 22 players. Note that for any given frame of the game, only some of the players are on screen and others are hidden off-screen and their information would have to be extrapolated by the computer vision system estimating the game's state from just the image, by learning about the rules and dynamics of the game of football.

\section{Self-Supervised Learning Methods} \label{sec:sslmethods} 

In this study we explore three types of self-supervised learning algorithms: a contrastive method (SimCLR), a contrastive method employing an online clustering approach (SwAV) and a non-contrastive method (BYOL). We choose to use three conceptually different SSL algorithms for two reasons. First, by using different algorithms we gain insights on the applicability of the SSL paradigm on game-state representation learning based only on games' pixel information. Second, we are able to investigate the degree to which different SSL approaches can efficiently learn informative game-state representations. We use the \emph{solo-learn} framework \cite{da2022solo} for the implementation of all SSL methods in our experiments. This section outlines the key elements of these three methods.

\subsection{SimCLR Approach}

SimCLR \cite{chen2020simple} is a contrastive method that learns image representations by pairwise comparison of similar and dissimilar images. In this method, we first take an image and apply certain content-preserving data augmentations such as scaling, rotation, jitter, etc. to obtain two different views of the same image (similar to \figurename{} \ref{fig:sslexample}). We define a loss such that the representation obtained from these two views of the same image lie close to each other in the feature space (positive pair), and at the same time, lie as far away as possible from representations of other images (negative pairs). %Thus, this method of learning representations from different views of the same image in comparison with other dissimilar images makes it a contrastive approach. 

\begin{figure*}[!tb]
\centerline{\includegraphics[width=\textwidth]{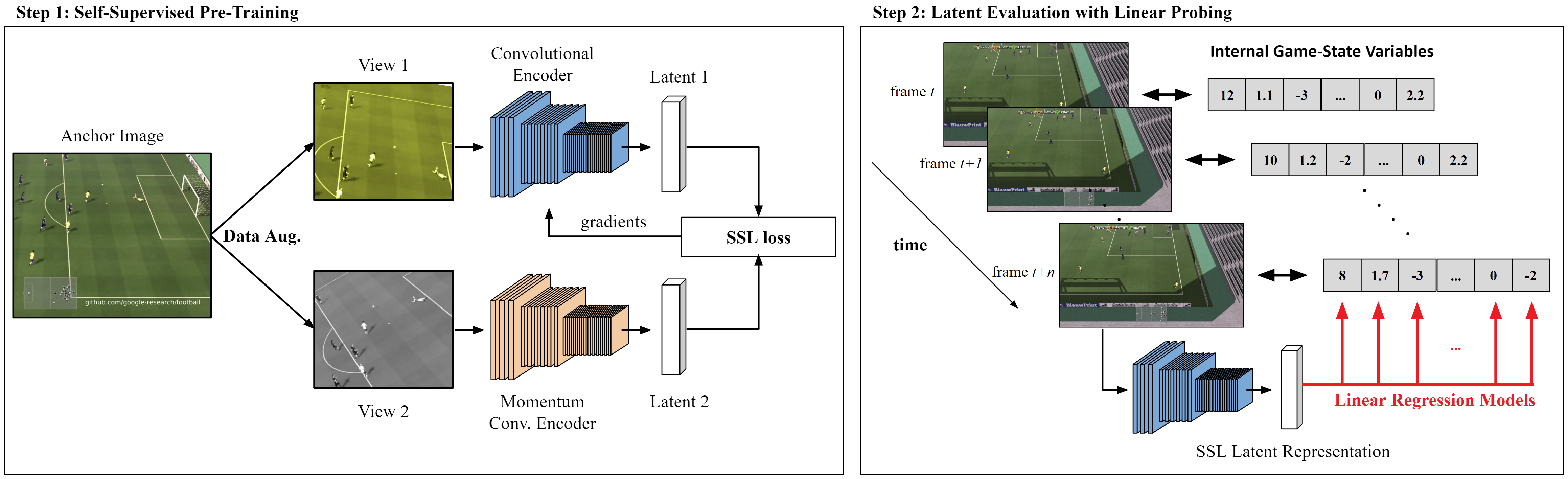}}
\caption{The two step process employed in this paper: we first pre-train the Convolutional Encoder using different SSL methods (left), then evaluate the learned representations with Linear Probing on the test set (right).}
\label{fig:method}
\end{figure*}

In terms of SSL in games, we can think of this method as trying to identify the important visual features of a gameplay frame that characterise the game's current state so that two views of the same game-state have similar representations, whereas different states have dissimilar representations. To achieve, however, good performance, especially for high-dimensional representations, this method requires a large number of negative samples. Therefore, during training, it requires a huge batch size to assure that the employed negative samples are representative of the entire dataset.

\subsection{BYOL Approach}

Bootstrap Your Own Latent (BYOL) is a non-contrastive approach to self-supervised learning suggested by Grill \emph{et al.} \cite{grill2020bootstrap}. This method does not require negative examples (dissimilar images), but only focuses on learning the similar representations of two views of the same image, hence the name. In absence of negatives, BYOL uses a stop-gradient method coupled with a Siamese-based network architecture to overcome the limitation of all representations collapsing to a constant (an undesired, trivial solution \cite{grill2020bootstrap}). This implies that the two views of the same image are propagated through two identical neural networks, but the gradients are passed through only one of the networks and the weights of the other network are updated as a moving-average of the first. This behavior simulates a ``memory'' mechanism activated during training, which indirectly provides the effect of using negative samples typically used in the contrastive learning paradigm. 

In terms of game-state representations, this method focuses solely on learning similarities between two views of the same gameplay frame. It makes it interesting to see how well this approach holds in distinguishing different game-states without the need of explicit negative examples.

\subsection{SwAV Approach}

Swapping Assignments between Views (SwAV) \cite{caron2020unsupervised} is another contrastive learning approach that attempts to address the requirement for large number of negative examples (and thus large batch sizes). SwAV uses an online clustering algorithm that maintains a codebook of clusters of different representations encountered during training. This codebook is formed by online clustering the derived representations based on the idea that differently augmented views of the same image (positive pair) should be clustered together, while representations corresponding to dissimilar images (negative pairs) should be assigned to different clusters. By imposing certain constraints such as assigning equally-distributed cluster labels to an input batch, this method prevents the collapsing representation problem \cite{caron2020unsupervised}. Finally, based on clusters' information, SwAV waives the need for large batch sizes. In theory, SwAV should be among the more practical and scalable methods for representation learning in games in terms of lower hardware-resource requirements for training. 

\section{Game Representation Learning}\label{sec:method}

This section presents our approach for training and evaluating the employed SSL algorithms on the problem of game-state representation learning based only on games' pixel information using the dataset described in Section \ref{sec:dataset}. To evaluate the quality of the SSL-based representations of gameplay frames' visual content, we use a ResNet50 \cite{he2016deep} model pretrained on the ImageNet dataset \cite{ILSVRC15} as a baseline.
%AL: shaving space
%ImageNet consists of 1.2M images belonging to 1K categories. Machine learning models pretrained on ImageNet can produce general purpose representations of images, and thus, this dataset has been widely used \cite{soares2019deep,peng2019novel,baveye2015deep} to pre-train models for transfer of learning purposes.

\subsection{Training}

%In this section we explain how we use the self-supervised learning algorithms to train a convolutional encoder able to capture the essential features describing the visual content of the gameplay frames. 
We use the ResNet50 architecture as our backbone model for transforming frames' pixels information to compressed yet informative game state representations. For all the employed games, the ResNet50 encoder receives as input RGB gameplay frames of dimension $224\times224\times3$ and compresses it to a representation vector of $2048$ real numbers. This encoder is trained with the three SSL algorithms described in Section \ref{sec:sslmethods} using the default hyper-parameters in \emph{solo-learn} \cite{da2022solo}. For each of the games, we use 50K images as training set and train the encoder for 50 epochs. The training step is visually presented in the left side of Fig. \ref{fig:method}.

% \begin{table*}[!tb]
% \begin{center}
% \caption{Minimum (min), average (avg) and maximum (max) $R^2$ correlation values between representations of images and the synchronized internal state variables across games. The best method (highest average $R^2$) per game is in bold.}
% \label{tab:results}
% \begin{tabular}{|l|lll|cll|lll|}
% \hline
% \multirow{2}{*}{\diagbox[height=2\line]{\textbf{Method}}{\textbf{Game}}} & 
% \multicolumn{3}{c|}{\textbf{ViZDoom}} & \multicolumn{3}{c|}{\textbf{CARLA}} & \multicolumn{3}{c|}{\textbf{GRFE}} \\ 
% & \textbf{min} & \textbf{avg} & \textbf{max} & \multicolumn{1}{l}{\textbf{min}} & \textbf{avg} & \textbf{max} & \textbf{min} & \textbf{avg} & \textbf{max} \\ \cline{1-10} 
% ImageNet & 0.42 & 0.68 & 0.78 & 0.52 & 0.59 & 0.63 & 0.08 & 0.11 & 0.19 \\
% SimCLR & 0.55 & 0.77 & 0.86 & 0.79 & 0.83 & 0.88 & 0.16 & 0.20 & 0.22 \\
% BYOL & \textbf{0.54} & \textbf{0.81} & \textbf{0.91} & \textbf{0.85} & \textbf{0.89} & \textbf{0.93} & 0.21 & 0.23 & 0.26 \\ 
% SwAV & 0.42 & 0.64 & 0.76 & 0.71 & 0.82 & 0.88 & \textbf{0.22} & \textbf{0.27} & \textbf{0.32} \\ \hline
% \end{tabular}

% \end{center}
% \end{table*}

\subsection{Evaluation}
\label{ssec:evaluation}

Once the backbone encoder has been trained, we evaluate the quality of the derived representations. To perform this, we prepare a separate evaluation dataset (detailed in Section \ref{sec:dataset}) consisting of images not seen during training, accompanied by the corresponding ground truth internal game state variables. The size of the evaluation dataset is different for each game because it is based on the appearance frequency of the internal game state variables. That means that when one or more game state variables are not present in a particular game state (e.g. no enemy is present on the screen in ViZDoom), we do not consider the frame that corresponds to that particular game state for evaluation purposes.

%As mentioned before, we evaluate the quality of the derived SSL-based representations with respect to the ground truth internal game state variables. 
Following the principles of \cite{anand2019unsupervised} the evaluation takes place by quantifying the capacity of a linear model to recover or predict the internal game state variables based on the derived SSL representations. This evaluation approach, called \emph{linear probing} \cite{anand2019unsupervised}, has been used to evaluate representations of Atari games. The internal state of Atari games is described via discrete variables, and thus linear probing in \cite{anand2019unsupervised} evaluated the capacity of a linear model to predict the class of the internal game state variables. In our study, however, the employed games' internal states are characterised by continuous variables. For this reason, instead of a linear classifier, we apply the linear probing technique with a linear regression model. 

% In particular.... \textcolor{red}{here you have to describe better the linear probing technique. You can mention what is the input of the linear regression model, what is its output and how this output is used to compute the correlation that you report.}

%GNY: continue reviewing from here....

In particular, for a game with $k$ internal state variables $V = \{v_1, v_2,.., v_k\}$, we train $k$ different linear regression models. For each of these models, the $d$-dimensional latent representations $z_d$ obtained from the backbone encoder are treated as the input or independent variables while the associated values in $V$ are treated as the dependent variables. With ResNet50 as the backbone in this paper, $d=2048$. For our dataset of $n$ pairs of images and each of the $k$ state variables, we fit the following least-squares linear regression model:
\begin{equation}
v_k^i = \beta_{k0} + \beta_{k1} z_1^i + ... + \beta_{k2048} z_{2048}^i + \epsilon^i\ \ \ \ \  \text{for}\ i = 1,...,n
\end{equation}
where $\beta$ are the coefficients and $\epsilon$ is the error. Here, we utilize the coefficient of determination (commonly known as $R^2$ correlation) of this model to quantify its performance. Higher values of $R^2$ indicate that the ResNet image encoder is better equipped to accurately extract the values of the internal state variables into the compressed representation, compared to those encoders with lower values of correlation. Thus, ideally, we want our $R^2$ values to be close to 1.0, so that any down-stream application that uses these representations works with the correctly interpreted state of the game from its image input.

At this point we should emphasize that the evaluation process relies only on linear probing models, as opposed to more complex nonlinear ones, since we want to evaluate the quality of the derived representations. This implies that the derived representations should be easily mapped to the internal game state variables via simple linear functions. In other words, we do not want to evaluate the prediction accuracy of a regression model, but the representation power of the SSL-based backbone encoder. The evaluation procedure is illustrated at the right image of Fig. \ref{fig:method}. In the following section we summarize the evaluation results obtained using $R^2$. %Chintan: can we give it a name and a variable so that we use it widely in graphs tables etc? Is it R2 correlation? Pearson correlation? What does Bengio use?
%Georgios: Bengio's paper has linear classifiers instead of regression, so they use F1 score of the classification numbers. I think coefficient of determination in linear regression is a basic enough concept in statistics to not elaborate on its math, just like Bengio don't go into any details of the F1 score. What do you think?$

\section{Results}\label{sec:results}
Table \ref{tab:results} presents the quality of representations produced by the baseline ImageNet and the three SSL methods in terms of $R^2$ correlation (see Section \ref{ssec:evaluation}) with the internal game state variables. For all games examined, SSL-based representations correlate better to the internal games state compared to the baseline method. Specifically, SSL algorithms yield average performance improvement, compared to the baseline model, of $19\%$, $51\%$ and $145\%$ for the ViZDoom, CARLA and GRFE games, respectively. In addition, SSL approaches managed to achieve a maximum $R^2$ value (best correlated internal state variable) higher than $0.9$ for two out of the three games.   

% \begin{figure*}[!tb]
% 	\begin{minipage}{0.5\linewidth}
% 		\centering
% 		\centerline{\includegraphics[width=1.0\linewidth]{figures/vizdoom_bar}}
% 	\end{minipage} 
% 	\begin{minipage}{0.5\linewidth}
% 		\centering
% 		\centerline{\includegraphics[width=1.0\linewidth]{figures/carla_bar}}
% 	\end{minipage}
% 	\begin{minipage}{1.0\linewidth}
% 		\centering
% 		\centerline{\includegraphics[width=1.0\linewidth]{figures/grfe_bar}}
% 	\end{minipage}
% 	\caption{Improvement of correlation values as a percentage over the ImageNet baseline model. Results are presented for all three games, their corresponding internal states considered and across the 3 SLL methods.}
% 	\label{fig:baseline_difference}
% \end{figure*}

\begin{figure*}[!tb]
    \centerline{\includegraphics[width=0.95\textwidth]{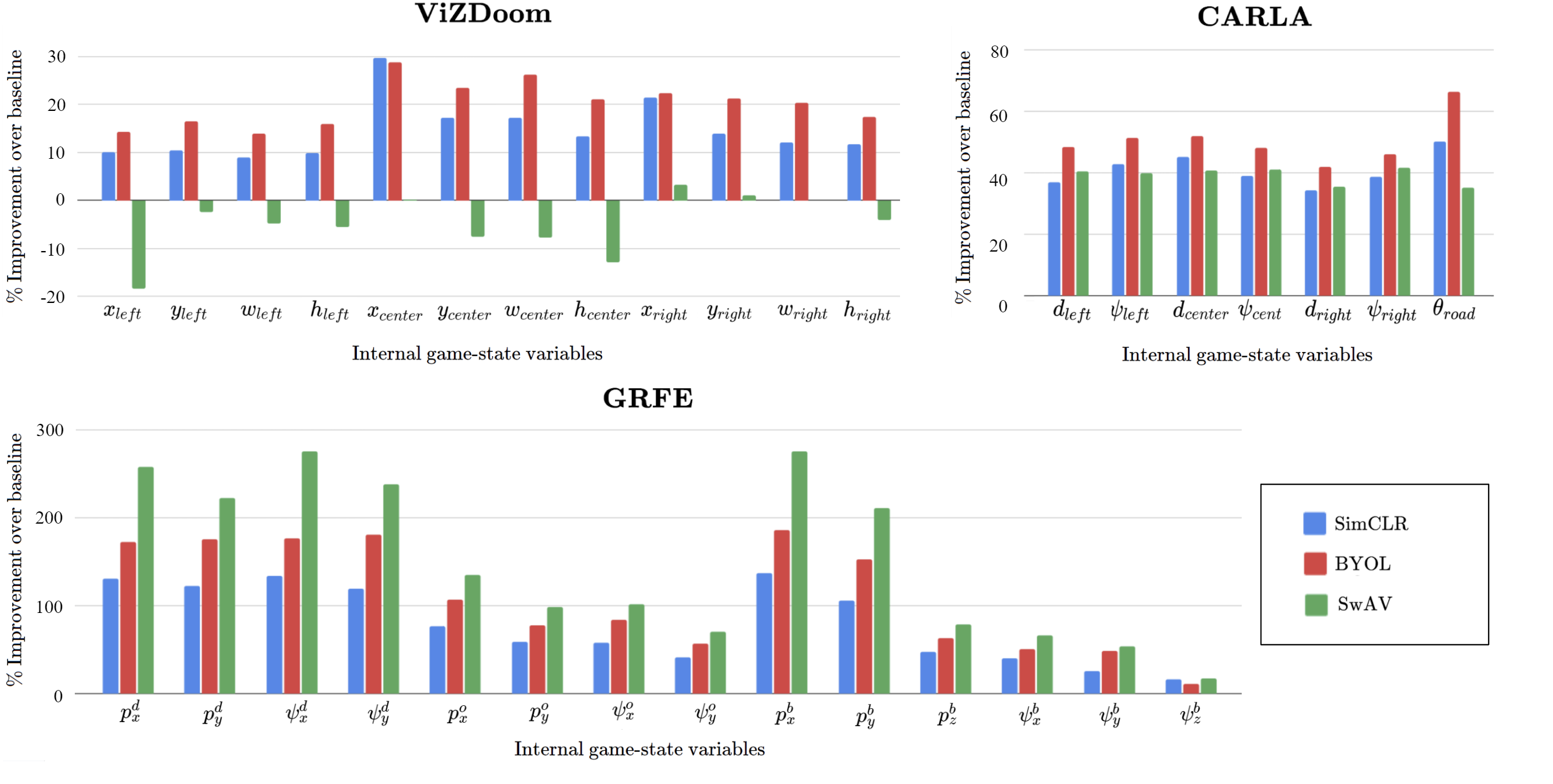}}
	\caption{Difference of $R^2$ correlation values as a percentage over the ImageNet baseline model. Results are presented for all three games, their corresponding internal states considered and across the 3 SSL methods.}
	\label{fig:baseline_difference}
\end{figure*}

Figure \ref{fig:baseline_difference} illustrates the percentage of improvement over baseline for the three SSL employed methods and for each of the internal game state variables. For all three games, BYOL and SimCLR appear to produce representations that are better correlated, compared to the baseline, to each one of their internal game state variables considered. The SwAV method, however, performs better than the baseline only for CARLA and GRFE. Nevertheless, it achieves the best performance on GRFE, which is the largest performance improvement across all games: three times higher $R^2$ values than the baseline. 

For the ViZDoom game, we observe that SSL representations yield higher improvements over the baseline for the internal state variables that correspond to the middle part of the gameplay frames. This behavior could perhaps be pertaining to the fact that the $(x,y)$ position values of the enemy on the left part of the screen lie towards the periphery of the image frame while that for the center and right parts of the screen lie closer to the center of the image, affecting the performance of the ResNet encoder owing to its convolutional architecture. %Chintan: Why is right closer to the center than left? This is not a valid argument in my opinion
% x,y are bottom left corner of bounding boxes seen in fig 2, i noticed these features used in the arnold paper and in the vizdoom api
% Chintan: Still I don't understand - sorry. This argument needs to be clearer/please revise the text and add details to support it

For the CARLA game the improvement across the internal state variables that correspond to the distance between the ego-vehicle and its surrounding vehicles, as well as to the direction of the surrounding vehicles presents very small fluctuations (between $35\%$ to $50\%$). We observe, however, a larger improvement for the road curvature internal state variable; i.e. $\sim$65\% of improvement with the BYOL method. %Chintan: it would be great to add a sentence here about the  "why" that happens for the curvature? Easy to predict? what makes BYOL predict curvature...?

\begin{table}[t]
\begin{center}
\caption{Minimum (min), average (avg) and maximum (max) $R^2$ correlation values between representations of images and the synchronized internal state variables across games. The best method (highest average $R^2$) per game is in bold.}
\label{tab:results}

\begin{tabular}{cl||l|l|l|l}
\hline\hline
\multicolumn{2}{l||}{} & \textbf{ImageNet} & \textbf{SimCLR} & \textbf{BYOL} & \textbf{SwAV} \\ \hline \hline
\multicolumn{1}{c}{\multirow{3}{*}{\textbf{ViZDoom}}} & \textbf{min} & \multicolumn{1}{c|}{0.42} & \multicolumn{1}{c|}{0.55} & \multicolumn{1}{c|}{\textbf{0.54}} & \multicolumn{1}{c}{0.42} \\
\multicolumn{1}{c}{} & \textbf{avg}  & \multicolumn{1}{c|}{0.68} & \multicolumn{1}{c|}{0.77} & \multicolumn{1}{c|}{\textbf{0.81}} & \multicolumn{1}{c}{0.64} \\
\multicolumn{1}{c}{} & \textbf{max}  & \multicolumn{1}{c|}{0.78} & \multicolumn{1}{c|}{0.86} & \multicolumn{1}{c|}{\textbf{0.91}} & \multicolumn{1}{c}{0.76} \\ \hline
\multicolumn{1}{c}{\multirow{3}{*}{\textbf{CARLA}}} & \textbf{min} & \multicolumn{1}{c|}{0.52} & \multicolumn{1}{c|}{0.79} & \multicolumn{1}{c|}{\textbf{0.85}} & \multicolumn{1}{c}{0.71} \\
\multicolumn{1}{c}{} & \textbf{avg} & \multicolumn{1}{c|}{0.59} & \multicolumn{1}{c|}{0.83} & \multicolumn{1}{c|}{\textbf{0.89}} & \multicolumn{1}{c}{0.82} \\
\multicolumn{1}{c}{} & \textbf{max} & \multicolumn{1}{c|}{0.63} & \multicolumn{1}{c|}{0.88} & \multicolumn{1}{c|}{\textbf{0.93}} & \multicolumn{1}{c}{0.88} \\ \hline
\multicolumn{1}{c}{\multirow{3}{*}{\textbf{GRFE}}} & \textbf{min} & \multicolumn{1}{c|}{0.08} & \multicolumn{1}{c|}{0.16} & \multicolumn{1}{c|}{0.21} & \multicolumn{1}{c}{\textbf{0.22}} \\
\multicolumn{1}{c}{} & \textbf{avg} &  \multicolumn{1}{c|}{0.11} & \multicolumn{1}{c|}{0.20} & \multicolumn{1}{c|}{0.23} & \multicolumn{1}{c}{\textbf{0.27}} \\
\multicolumn{1}{c}{} & \textbf{max} & \multicolumn{1}{c|}{0.19} & \multicolumn{1}{c|}{0.22} & \multicolumn{1}{c|}{0.26} & \multicolumn{1}{c}{\textbf{0.32}} \\ \hline \hline
\end{tabular}
\end{center}
\end{table}

Finally, for the GRFE game, since there is a large number of state variables (4 variables for each of the 22 players), we aggregate the correlation metrics as per the position of the players on the football pitch. For each of the 4 variables, we average the correlation figures for all the defensive players (1 goalkeeper and 4 defenders) of both teams. These 4 combined variables for positions and directions are labelled as $p_x^d$, $p_y^d$, $\psi_x^d$ and $\psi_y^d$ in \figurename{} \ref{fig:baseline_difference}. We repeat the same for the remaining 6 offensive players of both teams, labelled as $p_x^o$, $p_y^o$, $\psi_x^o$ and $\psi_y^o$. The 6 variables associated with the ball positions ($p^b$) and directions ($\psi^b$) are presented as-is. We observe higher improvements for the internal state variables of defensive players of both teams compared to their offensive players. 
We assume that this is because defensive players (especially the goalkeeper) move far less from their usual positions compared to offensive players. 
%We assume that these results derive from the fact that there is a larger variation in positioning of offensive players compared to defensive players due to the nature of the game of football. 
This behavior is embedded into the rules of football, and we expect that any SSL method employed in games will yield higher predictive capacity if it incorporates rules and game dynamics in its learning process. We notice that SwAV builds on its clustering capacity and achieves larger percentage improvement for variables corresponding to the defensive players and thus outperforms BYOL and SimCLR.

% Finally, for the GRFE game we observe higher baselines for the internal state variables of defensive players of both teams compared to their offensive players. We assume that this results derives from the fact that there is a larger variation in positioning of offensive players compared to defensive players due to the nature of the game of football. On that basis, we expect that any SSL method employed in games will yield higher predictive capacity if it incorporates rules and game dynamics in its learning process. We notice that SwAV takes advantage of this behavior and achieves larger percentage improvement for the variables corresponding to the defensive players and thus outperforms BYOL and SimCLR on average. %Chintan: this is good ...but it is still not clear what makes SwAV able to detect differences between offense and defense... its clustering feature maybe?

Based on the obtained results across games, we can conclude that SSL methods can produce better internal game state representations describing the games' internal state than models that are pretrained on huge datasets. The BYOL method seems the most robust of the three SSL methods tested in this work. It yields the best results for ViZDoom and CARLA games and the second best for the GRFE football game. The SwAV method achieves the best performance on the GRFE game; however, its behaviour seems to be highly affected by the game at hand as it performs worse than the baseline on the ViZDoom game. Finally, SimCLR produces consistent results across all games that outperform the baseline, although it is not the best performing for any of the three games examined. 

% \begin{table}[!h]
% \begin{center}
% \caption{Average Correlation observed between the representations of images and the synchronized internal state variables.}
% \label{tab:results}
% \begin{tabular}{|l|@{ }c@{ }|@{ }c@{ }|@{ }c@{ }|}
% \hline
% \diagbox[]{\textbf{Method}}{\textbf{Game}} & \textbf{\begin{tabular}[c]{@{}c@{}}ViZDoom\end{tabular}} & \textbf{\begin{tabular}[c]{@{}c@{}}CARLA\end{tabular}} & \textbf{\begin{tabular}[c]{@{}c@{}}GRFE\end{tabular}} \\ \hline
% ImageNet & \ 0.68 $\pm$ 0.09 \ & \ 0.59 $\pm$ 0.04 \ & \ 0.11 $\pm$ 0.03 \ \\ 
% SwAV &  \ 0.64 $\pm$ 0.10 \  & \ 0.82 $\pm$ 0.06 \ & \ 0.27 $\pm$ 0.03 \ \\ 
% BYOL & \ 0.81 $\pm$ 0.09 \ & \ 0.89 $\pm$ 0.03 \ & \ 0.23 $\pm$ 0.01 \  \\
% SimCLR & \ 0.77 $\pm$ 0.08 \ & \ 0.83 $\pm$ 0.04 \ & \ 0.20 $\pm$ 0.01 \  \\ \hline
% \end{tabular}

% \end{center}
% \end{table}

\section{Discussion}\label{sec:discussion}

The key takeaway from the experiments presented in this paper is that self-supervised learning, when applied directly to raw gameplay images, can derive game representations that are \textit{general}, as SSL manages to capture and correlate to key features of each game. Compared to the model pretrained on ImageNet, SSL is more efficient and robust across all three very different games tested: the games vary not only in terms of game mechanics but also in terms of image resolution, object sizes, and number of internal variables. Comparing the performance of the three SSL approaches examined we can conclude that the non-contrastive approach (BYOL) seems better suited for the task of state representation learning in complex 3D games, compared to the contrastive approaches. To solidify these conclusions, further investigation would be required with other contrastive and non-contrastive SSL methods such as Barlow Twins \cite{zbontar2021barlow}, SimSiam \cite{chen2020exploring}, VICReg \cite{bardes2021vicreg}, and DINO \cite{caron2021emerging}, as well as time-distributed SSL approaches such as ST-DIM \cite{anand2019unsupervised}. Beyond testing our hypothesis across more SSL algorithms, we plan to test the robustness of the proposed method across a larger variety of games and game-genres with dissimilar types of graphics, aesthetics, and rules.

In terms of applications of self-supervised learning in AI and games research \cite{yannakakis2018artificial}, we recommend that when using convolutional networks for processing visual input one should take advantage of pretraining the network using SSL. Since the optimization criteria in SSL methods are independent of the end-task, such methods can provide more general-purpose representations that can be used for a multitude of tasks. This covers a wide range of applications such as training and testing game-playing bots, deep reinforcement learning, procedural content generation, affective computing, player experience modeling, and generative modeling, which all can make use of the general features learned by these SSL frameworks. This should not only improve their performance by providing more meaningful and informative input from the game, but also help improve efficiency in learning the indicated task in the game. In addition, the training time of any given learning task is significantly reduced since the processing and compression of raw pixels is handled by the general-purpose SSL pre-trained models. As a result one can simply focus on optimizing the objective of the learning task at hand.

\section{Conclusion}\label{sec:conclusion}

In this study we demonstrated that self-supervised learning methods can be used for learning highly informative, descriptive and general-purpose representations from RGB images of games. In particular, we presented a new dataset of three dissimilar games in terms of genre, footage resolution and key object sizes that appear on screen: VizDoom (clone of \emph{Doom} first-person shooter), CARLA (racing game simulator), and GRFE (football game simulator). The dataset contains the internal state of the game---a vector of critical features about each game---with the corresponding RGB frames seen in the game's renderer. To test our hypothesis that SSL derives more descriptive and hence general representations of games, we employ three representative SSL methods and attempt to predict the internal state values of the three games from their RGB frames. Our results suggest that SSL-based representations are more powerful than general-purpose pre-trained models at correctly extracting the internal game states from images. This comes without any cost of labor since the SSL methods are trained on just images and no special annotations or manual labelling is required. Our key findings suggest that SSL is not only a practical but a highly recommended approach for deriving general-purpose and meaningful compressed representations for dissimilar AI task within games: from gameplaying and testing agents, and generative/creative AI systems all the way to player modeling tasks.

\bibliography{ssl_games}

\end{document}